\documentclass[11pt]{article}

\usepackage[preprint]{acl}

\usepackage{times}
\usepackage{latexsym}

\usepackage[T1]{fontenc}

\usepackage[utf8]{inputenc}

\usepackage{microtype}

\usepackage{inconsolata}

\usepackage{graphicx}
\usepackage{graphicx}
\usepackage{amsmath}
\usepackage{wrapfig}
\usepackage{subfigure}
\usepackage{booktabs}

\title{Optimal Expert-Attention Allocation in Mixture-of-Experts: A Scalable Law for Dynamic Model Design}

\author{
Junzhuo Li\textsuperscript{\dag}\textsuperscript{\ddag}, 
Peijie Jiang\textsuperscript{\P}, 
Changxin Tian\textsuperscript{\P}, 
Jia Liu\textsuperscript{\P}, \\
\textbf{Zhiqiang Zhang}\textsuperscript{\P}, 
\and 
\textbf{Xuming Hu}\textsuperscript{\dag}\textsuperscript{\ddag}\thanks{Corresponding author} \\
\textsuperscript{\dag}The Hong Kong University of Science and Technology (Guangzhou) \\
\textsuperscript{\ddag}The Hong Kong University of Science and Technology \\
\textsuperscript{\P}Ant Group \\
\texttt{jz.li@connect.hkust-gz.edu.cn} \\
\texttt{xuminghu@hkust-gz.edu.cn}
}

\begin{document}
\maketitle
\begin{abstract}
This paper presents a novel extension of neural scaling laws to Mixture-of-Experts (MoE) models, focusing on the optimal allocation of compute between expert and attention sub-layers. As MoE architectures have emerged as an efficient method for scaling model capacity without proportionally increasing computation, determining the optimal expert-attention compute ratio becomes critical. We define the ratio $r$ as the fraction of total FLOPs per token dedicated to the expert layers versus the attention layers, and explore how this ratio interacts with the overall compute budget and model sparsity. Through extensive experiments with GPT-style MoE Transformers, we empirically find that the optimal ratio $r^*$ follows a power-law relationship with total compute and varies with sparsity. Our analysis leads to an explicit formula for $r^*$, enabling precise control over the expert-attention compute allocation. We generalize the Chinchilla scaling law by incorporating this architectural parameter, providing a new framework for tuning MoE models beyond size and data. Our findings offer practical guidelines for designing efficient MoE models, optimizing performance while respecting fixed compute budgets.
\end{abstract}

\section{Introduction}
Modern large language models (LLMs) are increasingly trained under strict and explicit compute budgets. 
In industrial settings, model developers must operate within fixed GPU resources, training windows, and inference throughput constraints, making efficient use of computation a primary design objective rather than an afterthought . 
Mixture-of-Experts (MoE) \citep{shazeer2017, fedus-2021-switch, lepikhin2021gshard, albert-2024-mixtral, dai-etal-2024-deepseekmoe, muennighoff2024olmoe, qwen_moe} architectures have emerged as a practical solution in this regime, enabling substantial parameter growth while keeping per-token computation nearly constant through sparse expert activation.


However, adopting MoE architectures introduces new architectural decisions that are largely absent in dense Transformers \citep{vaswani2017attention}.
Beyond choosing the number of experts or the activation sparsity, practitioners must implicitly decide how computation is distributed \emph{inside} the model.
In particular, how much of the available compute should be allocated to attention layers versus expert (feed-forward) layers remains an open question.
In practice, this allocation is often inherited from dense Transformer designs or tuned heuristically, despite the fact that expert layers can dominate the compute budget in large-scale MoE models.


From a systems perspective, this raises a fundamental yet underexplored question:
\emph{given a fixed training compute budget, what is the optimal way to allocate computation between attention and expert components in an MoE Transformer?}
We characterize this trade-off using the FLOPs ratio
$r = C_E / C_A$, which directly controls where the majority of per-token computation is spent.
Unlike traditional hyperparameters, $r$ governs a first-order resource allocation decision and can substantially affect model performance at scale.

Existing neural scaling laws \citep{kaplan2020scaling, henighan2020scaling, hoffmann2022training, chowdhery2023palm} provide powerful guidance for allocating compute across model size and data.
The Chinchilla scaling law, for example, prescribes an optimal balance between parameters and training tokens under a fixed compute budget.
Recent extensions to MoE models further incorporate sparsity and expert count into this framework \citep{ludziejewski2024scaling, abnar2025parameters, wang-etal-2024-scaling}.
However, these approaches implicitly assume a fixed internal compute allocation, leaving the expert--attention trade-off unmodeled.
As a result, current scaling laws offer limited guidance on how MoE architectures themselves should evolve with increasing compute.

In this work, we show that the optimal expert--attention compute allocation is not fixed, but instead follows a predictable scaling behavior.
Through controlled experiments across multiple model scales and sparsity regimes, we demonstrate that the optimal FLOPs ratio $r^*$ increases as a power law of total training compute, with scaling coefficients that depend systematically on sparsity.
Building on this observation, we incorporate the expert--attention trade-off into a unified scaling law, extending the Chinchilla framework to account for internal architectural allocation.
Our results provide practical, compute-aware guidelines for designing MoE models that achieve better performance under fixed resource budgets.

\section{Theoretical Motivation}


This section provides a minimal theoretical rationale for why the optimal compute allocation between attention and expert components in MoE models should vary with scale.
Rather than proposing a detailed mechanistic model, our goal is to capture the essential trade-off underlying expert--attention compute allocation and to motivate the empirical scaling behavior observed in later sections.






\subsection{Compute Allocation in MoE Transformers}

In an MoE Transformer, the dominant per-token computation arises from two sources: self-attention and expert (feed-forward) layers.
Let $C_A$ denote the FLOPs consumed by attention and $C_E$ the FLOPs consumed by expert layers.
Under a fixed per-token compute budget $C = C_A + C_E$, their relative allocation is fully characterized by the FLOPs ratio
$r = C_E / C_A$.

In addition to compute allocation, MoE models are defined by their activation sparsity.
We characterize sparsity by the fraction of inactive experts,
$S = (E - E_{\text{act}})/E,$
where $E$ is the total number of experts and $E_{\text{act}}$ is the number of experts activated per token.
Sparsity controls how expert capacity is utilized: lower sparsity implies that more experts contribute computation per token, while higher sparsity concentrates compute on fewer experts.

Unlike dense Transformers, MoE architectures expose both $r$ and $S$ as explicit design choices.
Fixing either implicitly constrains the effectiveness of the other.
As a result, optimal compute allocation cannot be determined independently of sparsity.





\subsection{Diminishing Returns under Sparse Expert Activation}

Allocating additional compute to either attention or expert layers yields diminishing returns.
However, in MoE models, the rate of diminishing returns for expert computation depends critically on sparsity.
When sparsity is low (i.e., more experts are activated), additional expert compute can be distributed across multiple specialized subnetworks, leading to higher marginal gains.
In contrast, under high sparsity, expert computation is concentrated on a small set of experts, and additional compute is more likely to saturate existing representations.

This asymmetry implies that the effective elasticity of expert compute is sparsity-dependent.
While attention computation primarily benefits global token interaction modeling and is largely insensitive to sparsity, expert computation derives its benefit from both capacity and diversity, the latter being directly controlled by $S$.

\begin{figure*}[t]
    \centering
    \subfigure{\includegraphics[width=0.48\linewidth]{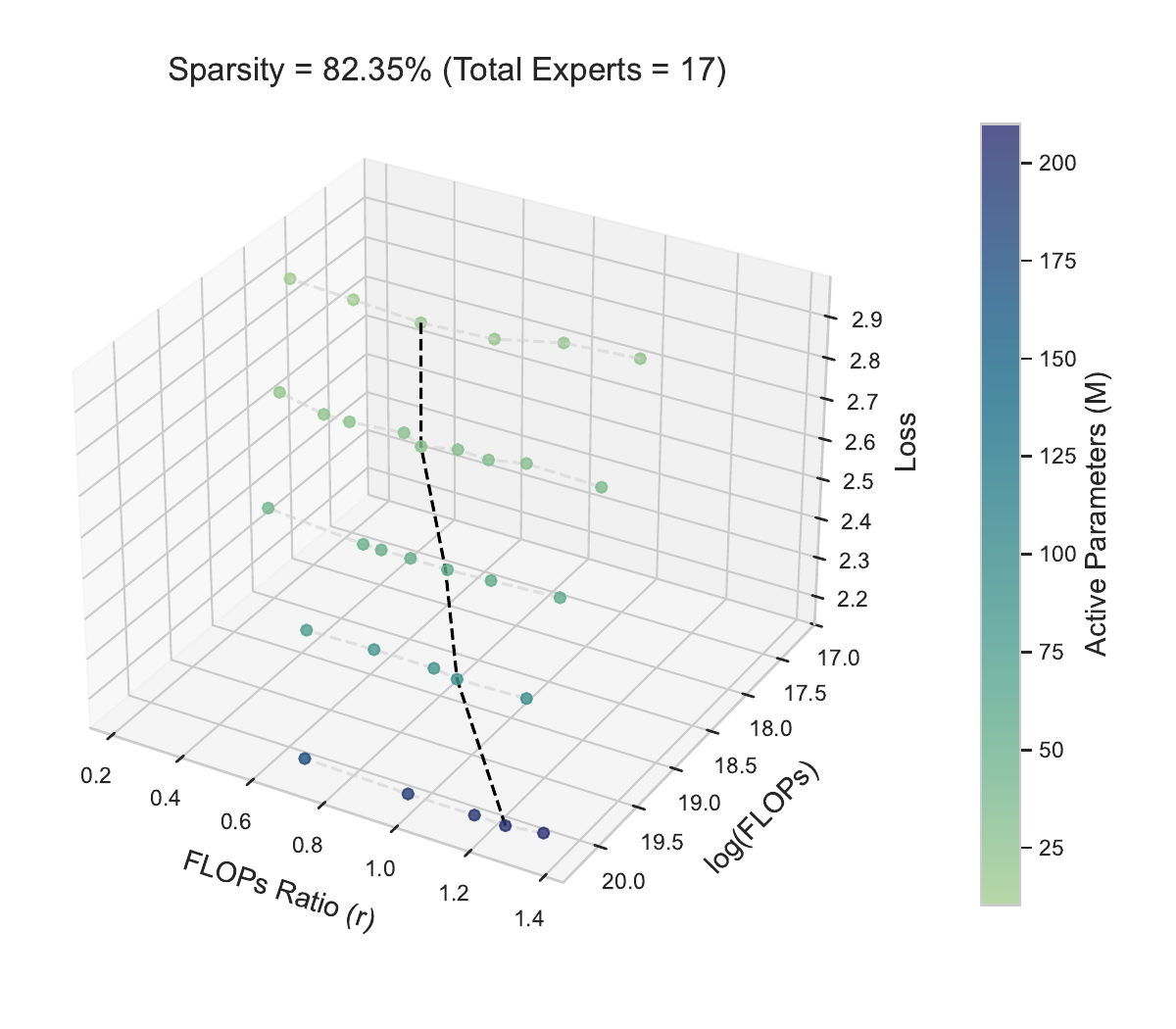}}
    \subfigure{\includegraphics[width=0.48\linewidth]{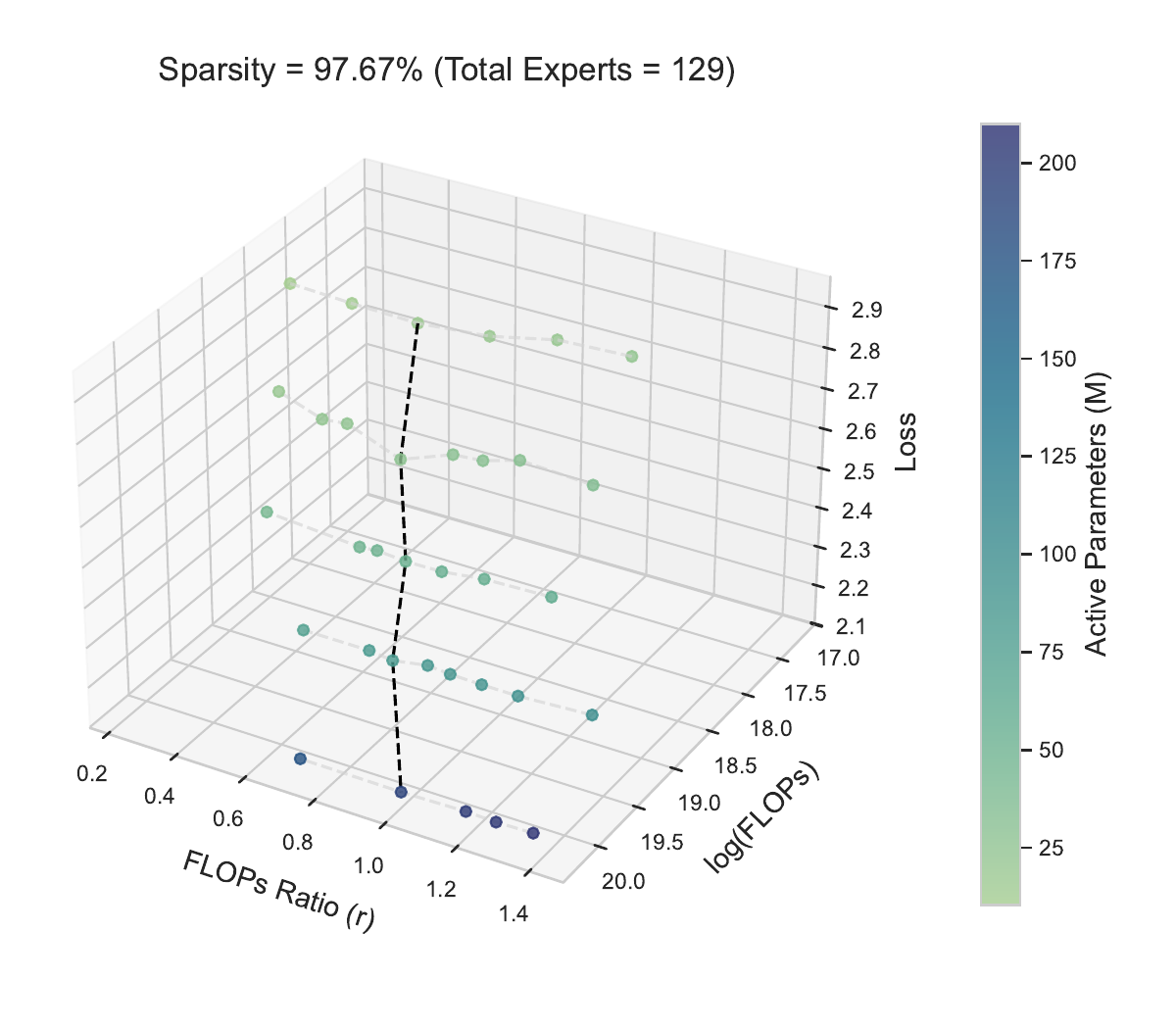}}
    \caption{Loss as a function of FLOPs ratio and total compute. Black dashed lines trace optimal $r^*$. Color indicates active parameters. Low-sparsity models (left) favor higher $r^*$ at scale.}
    \label{fig:flops_ratio_loss}
\end{figure*}

\subsection{Implications for Optimal Allocation}

Under a fixed compute budget, optimal performance is achieved by balancing the marginal utility of attention and expert computation.
Because sparsity modulates the effectiveness of expert compute, the optimal FLOPs ratio $r^*$ must depend on both total compute $C$ and sparsity $S$.
A minimal consequence of this interaction is that $r^*$ is not constant, but instead follows a scale-dependent law of the form
\begin{equation}
    r^*(C, S) = \alpha(S) C^{\beta(S)},
\end{equation}
where the sparsity-dependent coefficients reflect how expert utilization changes under different activation regimes.

This formulation yields a clear, testable prediction:
models with lower sparsity should benefit more from allocating additional compute to expert layers as scale increases, while highly sparse models should favor relatively greater attention capacity.
In the following sections, we empirically validate this prediction and quantify how sparsity shapes the scaling behavior of $r^*$.

\section{Empirical Scaling Behavior of the Optimal Compute Allocation}

Motivated by the sparsity-aware allocation rationale developed in Section~2, we now empirically investigate how the optimal expert--attention FLOPs ratio emerges in practice.
Rather than evaluating a particular MoE architecture, our focus is on characterizing the behavior of the optimal allocation itself as a function of total training compute and sparsity.
Specifically, we aim to answer three questions:
(i) whether a well-defined optimal ratio $r^*$ exists under fixed compute and sparsity,
(ii) how this ratio evolves as the compute budget increases, and
(iii) whether sparsity systematically modulates this evolution.

To this end, we perform controlled sweeps over the FLOPs ratio $r$ while holding the per-token compute budget fixed, across multiple model scales and sparsity regimes.
This experimental design isolates the effect of internal compute allocation from confounding factors such as parameter count or data scale.
We show that the optimal ratio $r^*$ is stable, scale-dependent, and exhibits consistent trends that form the empirical basis for the scaling laws developed in the following sections.

\begin{figure*}[t]
    \centering
    \subfigure[Optimal ratio vs.\ $C$]{\label{subfig:c2r}
        \includegraphics[width=0.31\linewidth]{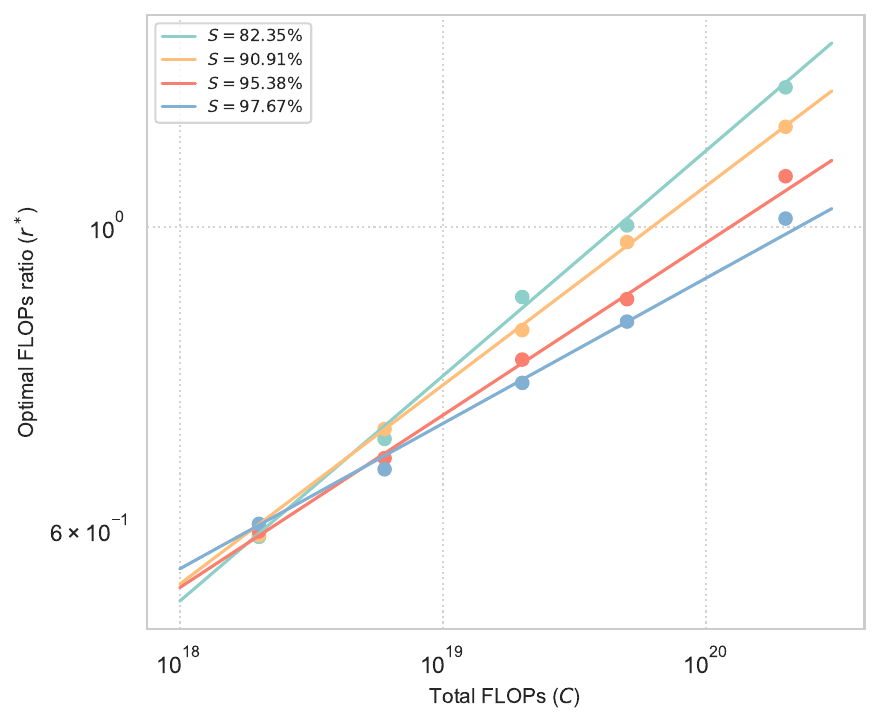}
    }
    \subfigure[Coefficient $\alpha_r$ vs.\ $1-S$]{\label{subfig:s2a}
        \includegraphics[width=0.31\linewidth]{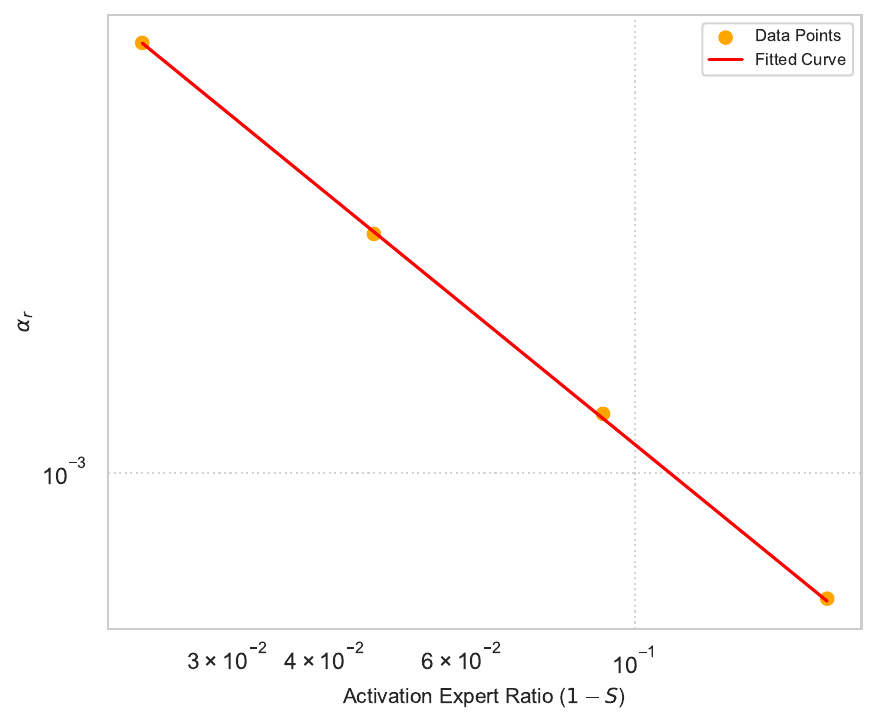}
    }
    \subfigure[Exponent $\beta_r$ vs.\ $1-S$]{\label{subfig:s2b}
        \includegraphics[width=0.31\linewidth]{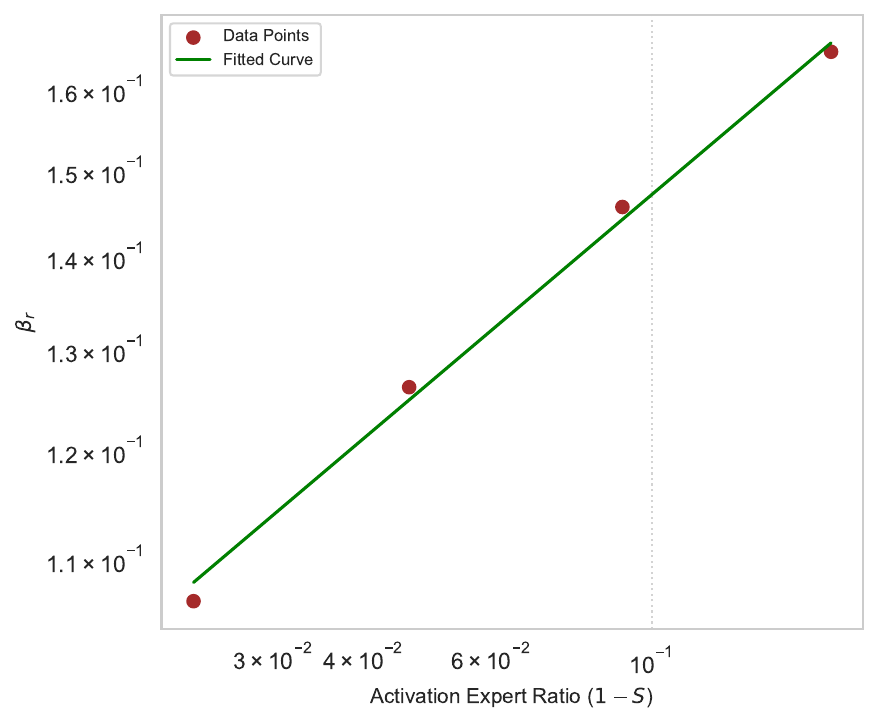}
    }
    \caption{
    {\bf (a)} Relationship between the optimal FLOPs ratio $r^*$ and total per‐token compute $C$, fitted with the power law $r^* = \alpha_r C^{\beta_r}$.  
    {\bf (b)} Dependence of the fitted coefficient $\alpha_r$ on the fraction of activated experts $(1-S)$, showing a clear power‐law trend.  
    {\bf (c)} Dependence of the fitted exponent $\beta_r$ on $(1-S)$, also following a power‐law.  
    All axes are plotted on log–log scales.
    }
    \label{fig:flpos_optimal_ratio}
\end{figure*}

\subsection{Existence of a Scale-Dependent Optimal Ratio}

We first examine whether a well-defined optimal FLOPs ratio $r^*$ exists under fixed training compute and sparsity.
Figure~\ref{fig:flops_ratio_loss} visualizes the training loss as a function of the FLOPs ratio $r$ and total compute $C$ for two representative sparsity levels.
Across all compute budgets, the loss surface exhibits a clear and smooth minimum along the $r$ dimension, forming a pronounced valley rather than a flat or noisy region.
This observation indicates that $r^*$ is a stable and well-defined quantity, rather than an artifact of random variation or overfitting.

As the total compute increases, the location of this minimum shifts systematically toward larger values of $r$, indicating that allocating proportionally more compute to expert layers becomes increasingly beneficial at scale.
Importantly, this shift is monotonic and smooth, suggesting a structured dependence on compute rather than isolated local effects.
These trends are consistently observed across model sizes and training runs, confirming that the optimal ratio $r^*$ is a robust property of the training regime.

Comparing the two sparsity settings in Figure~\ref{fig:flops_ratio_loss} further reveals that sparsity strongly influences how rapidly the optimal ratio evolves with compute.
In lower-sparsity models (left), where more experts are activated per token, the optimal $r^*$ increases more steeply as compute grows.
In contrast, under higher sparsity (right), the shift in $r^*$ is more gradual, reflecting a reduced marginal benefit from additional expert computation.
This qualitative difference provides direct empirical evidence that sparsity modulates the scale-dependent effectiveness of expert computation.

These results establish that a stable, scale-dependent optimum exists and that sparsity systematically reshapes its evolution.

\subsection{Scaling of the Optimal Ratio with Compute}

We next examine how the optimal FLOPs ratio $r^*$ evolves with increasing training compute.
For each fixed sparsity level, we extract $r^*$ from the loss minima in Figure~\ref{fig:flops_ratio_loss} and plot it against the total per-token compute $C$.
As shown in Figure~\ref{fig:flpos_optimal_ratio}(a), $r^*$ increases monotonically with compute across all sparsity regimes and follows an approximately linear trend on log--log scales.

This behavior is well captured by a power-law relationship,
\[
r^* = \alpha_r C^{\beta_r},
\]
which provides an accurate description of the observed data across model scales.
These results establish that the optimal expert--attention allocation is itself a scale-dependent quantity rather than a fixed architectural constant.

\subsection{Sparsity-Dependent Scaling Coefficients}

We now analyze how sparsity modulates the scaling behavior of the optimal FLOPs ratio.
Figures~\ref{subfig:s2a} and \ref{subfig:s2b} show the dependence of the fitted coefficients $\alpha_r$ and $\beta_r$ on the fraction of activated experts $(1-S)$.
Both parameters vary systematically with sparsity: $\alpha_r$ decreases while $\beta_r$ increases as more experts are activated.

These trends indicate that sparsity affects not only the magnitude of the optimal ratio, but also its rate of growth with compute.
Lower-sparsity models exhibit a steeper increase in $r^*$ as scale increases, whereas under high sparsity the growth is more gradual.
This confirms that sparsity reshapes the scaling regime of expert--attention allocation rather than acting as a simple constant offset.

Empirically, we find that the scaling coefficients follow simple power-law relationships with sparsity:
\begin{equation*}
    \begin{aligned}
        &\alpha_r = 6.7 \times 10^{-5} (1-S)^{-1.23}, \\
        &\beta_r = 0.24 (1-S)^{0.21}.
    \end{aligned}
\end{equation*}
These expressions provide a closed-form parameterization of the optimal allocation across sparsity regimes.

\subsection{Summary: An Empirical Law for Optimal Allocation}

The results in this section establish that the optimal expert--attention FLOPs ratio is a well-defined and scale-dependent quantity.
Across a wide range of model scales and sparsity regimes, the optimal ratio $r^*$ follows a simple power-law relationship with total compute, with scaling coefficients that vary systematically with sparsity.
Together, these findings yield an empirical allocation law of the form $r^*(C, S) = \alpha_r(S)\, C^{\beta_r(S)}$, which compactly captures how internal compute allocation should co-evolve with model scale and sparsity.
In the next section, we incorporate this allocation law into a unified scaling framework to quantify its impact on training loss.




\begin{figure*}[t]
    \centering
    \subfigure[Fit on data used to estimate coefficients.]{
        \includegraphics[width=0.48\linewidth]{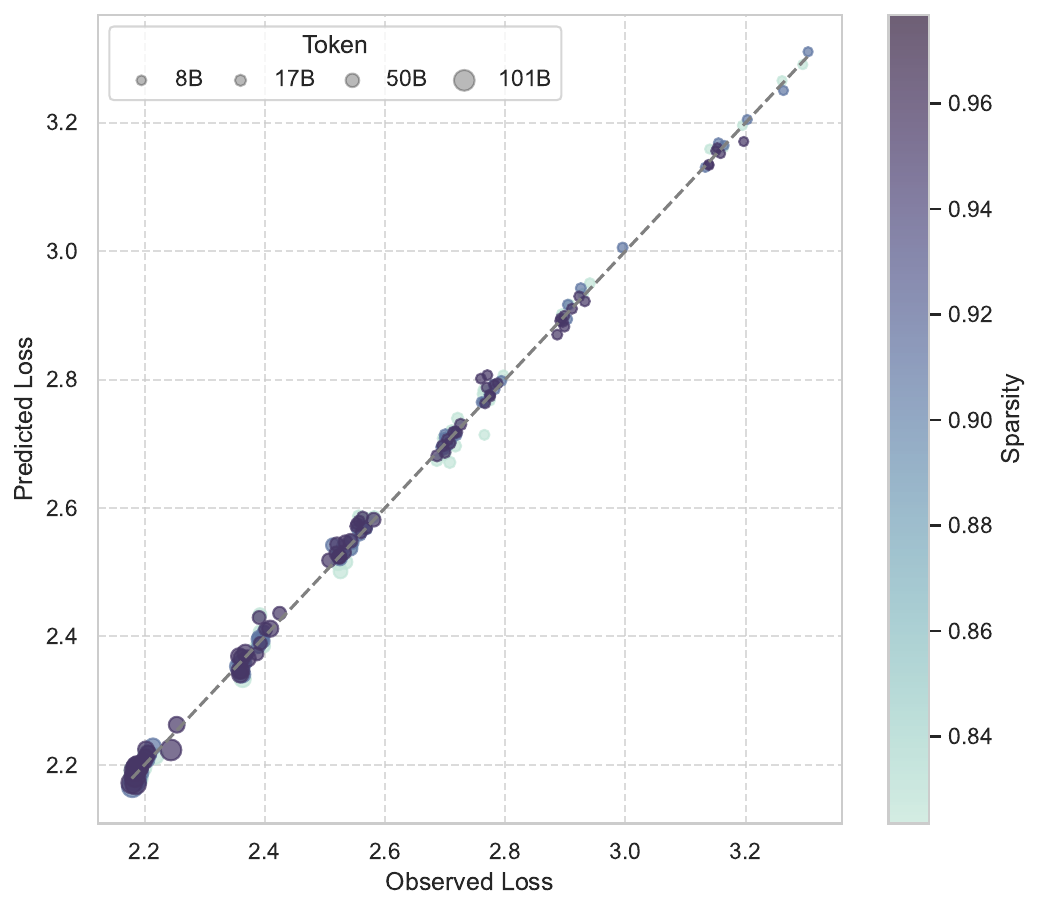}
    }
    \subfigure[Validating scaling law on held-out dataset.]{
        \includegraphics[width=0.48\linewidth]{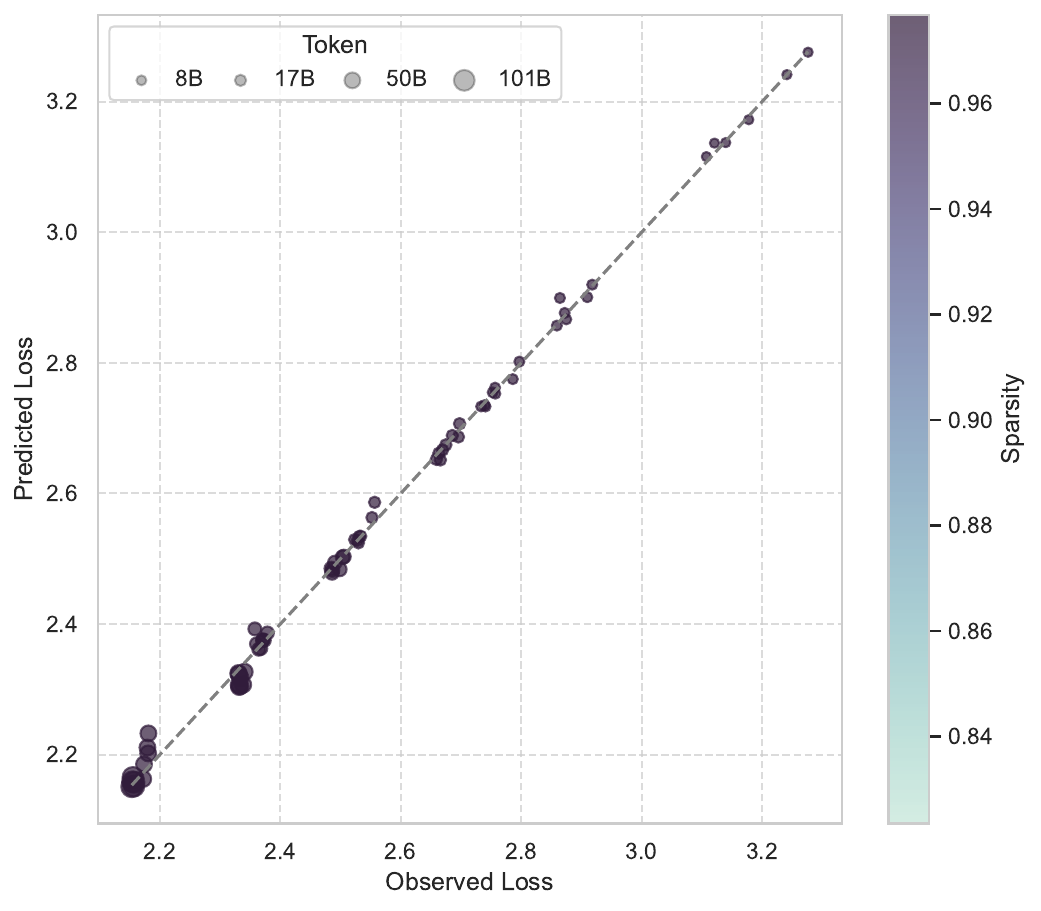}
    }
    \caption{Scaling law fit on data obtained from training compute-optimal models. {\bf (a)} shows the fit on the data used to estimate the coefficients for Eq.~\ref{eq:final}, {\bf (b)} validates these coefficients on a held-out dataset. All data points with S = 97.67\% were excluded from the fitting process for out-of-sample validation.
    }
    \label{fig:predicted_vs_observed}
\end{figure*}

\section{Scaling Laws with Expert--Attention Trade-offs}

Section~3 establishes that the optimal expert--attention FLOPs ratio $r^*$ follows a scale- and sparsity-dependent allocation law.
We now incorporate this allocation law into a loss-level scaling framework to quantify the impact of internal compute misallocation.
By explicitly modeling deviations from the optimal ratio, we extend conventional scaling laws to account for architectural allocation effects under fixed compute budgets.

\label{sec:chinchilla}
\subsection{From Allocation Law to Loss Scaling}

The empirical results in Section~3 indicate that performance under a fixed compute budget depends not only on how much compute is used, but also on how it is allocated between attention and expert components.
Conventional scaling laws implicitly assume a fixed internal allocation, treating models at the same compute budget as equivalent.
Our results show that this assumption leads to systematic performance degradation when compute is misallocated.

To capture this effect, we extend existing scaling laws with two additional terms that penalize deviations from the optimal FLOPs ratio and excessive expert allocation.
These terms do not introduce new resources; instead, they quantify the inefficiency arising from suboptimal use of a fixed compute budget.

To instantiate this formulation, we adopt the following extended scaling law:
\begin{equation}
\begin{aligned}
\mathcal{L} =\;& \frac{a}{N^\alpha}
+ \frac{b}{D^\beta} \\
&+ c\cdot\frac{e^{R}(1-S)^\gamma}{N^\lambda}
+ d \cdot \frac{r}{r+1}
+ \tau,
\end{aligned}
\label{eq:final}
\end{equation}
where $a, b, c, d, \alpha, \beta, \lambda, \gamma, \tau$ are fitted parameters.
In the next subsection, we evaluate the predictive accuracy of this formulation.

\subsection{Empirical Validation of the Extended Scaling Law}

We evaluate the predictive accuracy of the extended scaling law in Equation~\ref{eq:final}, which augments conventional loss scaling with explicit penalties for expert--attention misallocation.
Figure~\ref{fig:predicted_vs_observed}(a) shows strong agreement between predicted and observed loss on the data used for coefficient estimation.
More importantly, Figure~\ref{fig:predicted_vs_observed}(b) validates the same coefficients on a held-out sparsity level that was entirely excluded from fitting, demonstrating robust out-of-sample generalization.

\begin{figure}
    \centering
    \includegraphics[width=0.48\textwidth]{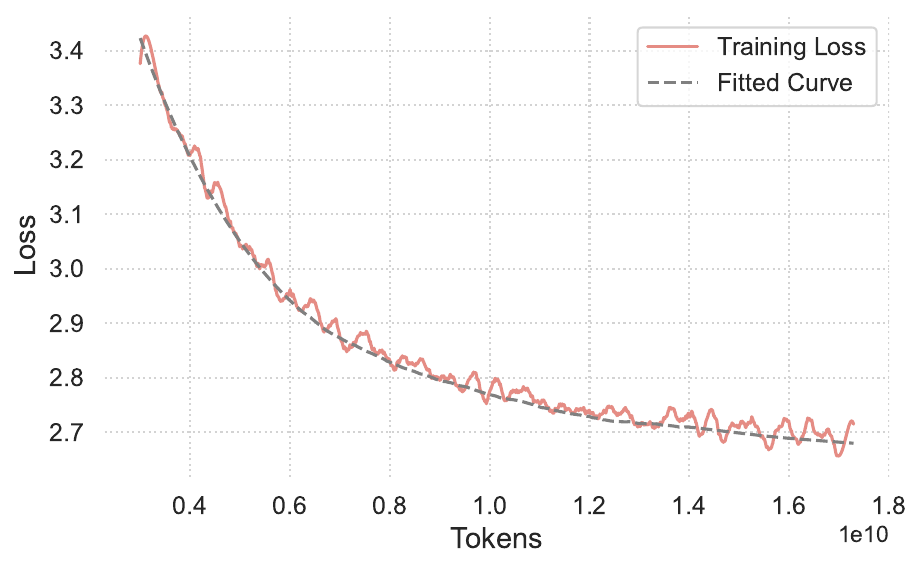}
    \caption{Comparison between the actual training curve of a model with 30M activation parameters and 550M total parameters (with 95.38\% sparsity) and the fitted curve derived from Equation~\ref{eq:final}.}
    \label{fig:loss_vs_fitted_curve}
    \vspace{-10pt}
\end{figure}

To assess whether the model captures training dynamics beyond final loss values, Figure~\ref{fig:loss_vs_fitted_curve} compares predicted and observed loss trajectories for a representative MoE model.
The close alignment indicates that the extended scaling law provides a consistent description of loss evolution under fixed compute allocation.
Estimated coefficients are reported in Table~\ref{tab:cofficient}.

\begin{table*}[t]
  
  \centering
  \resizebox{0.95\textwidth}{!}{
  \begin{tabular}{lccccccccc}
    \toprule
    Coefficient & $\alpha$ & $\beta $ & $\lambda$ & $\gamma $ & $\tau$ & a & b & c & d \\
    \midrule
    Estimate &0.6288 & 0.0453 & 0.4228 & 0.0431 & 13.7354 & 15.12 & 18.62 & 39.55 & 0.0499\\
    \bottomrule
  \end{tabular}}
  \caption{ Estimated values for coefficients in Equation~\ref{eq:final}.}
  \label{tab:cofficient}
\end{table*}

\subsection{Practical Implications under Fixed Compute Budgets}

The extended scaling analysis implies that optimal MoE design requires internal compute allocation to co-scale with both total compute and sparsity.
Maintaining a fixed expert--attention ratio across model scales leads to systematic inefficiencies, either underutilizing expert capacity at large scale or overspending compute on experts under high sparsity.

In practice, given a target compute budget $C$ and sparsity level $S$, the empirical allocation law $r^*(C,S)$ provides a direct guideline for selecting the appropriate expert capacity.
This enables architecture-level optimization under fixed resource constraints, ensuring that additional compute is translated into effective modeling capacity rather than wasted through suboptimal allocation.

\section{Related Work}

\paragraph{Neural Scaling Laws.}
Prior work has established that language model performance follows power-law relationships with model size, data, and compute \citep{kaplan2020scaling, hoffmann2022training}.
These scaling laws provide principled guidance for allocating total training resources, but typically assume fixed internal architectural configurations.

\paragraph{Scaling Laws for MoE Models.}
Mixture-of-Experts (MoE) architectures \citep{fedus-2021-switch, lepikhin2021gshard, dai-etal-2024-deepseekmoe} decouple parameter count from per-token compute through sparse expert activation.
Recent studies analyze how performance scales with total parameters, expert count, granularity, and sparsity \citep{clark2022unified, ludziejewski2024scaling, wang-etal-2024-scaling, abnar2025parameters}.
However, these works largely treat internal compute allocation as fixed and do not explicitly model how compute should be distributed between attention and expert components.

\paragraph{Our Contribution.}
In contrast, we treat internal compute allocation as a first-order scaling variable.
Rather than optimizing only total capacity or sparsity, we model how expert--attention FLOPs allocation should co-scale with compute and sparsity.
This perspective extends existing scaling frameworks by explicitly accounting for architectural compute misallocation and its impact on training loss.

\section{Discussion}

\subsection{Compute Allocation as a Scaling Variable}

Our results show that expert--attention compute allocation in MoE models is a scale- and sparsity-dependent variable rather than a fixed hyperparameter.
The optimal FLOPs ratio forms an additional scaling dimension that must co-evolve with total compute; otherwise, misallocation leads to measurable performance loss.
Sparsity further reshapes this allocation regime, with lower sparsity favoring expert-heavy configurations and higher sparsity favoring greater attention capacity.
Architecture-level compute distribution should therefore be optimized jointly with sparsity and training scale.

\subsection{Limitations}

Our analysis is restricted to autoregressive language modeling with fixed sparsity and does not account for multimodal tasks, adaptive routing, or hardware-level communication costs, which remain important directions for future work.


\section{Conclusion}

In this paper, we establish that expert--attention compute allocation is a critical and previously under-modeled dimension of MoE scaling.
The optimal FLOPs ratio follows a predictable law governed by total compute and sparsity, and deviations from this optimum reduce efficiency.
By integrating allocation-aware modeling into a Chinchilla-style framework, we move MoE design from static heuristics toward compute-optimal co-scaling, providing practical guidance for resource-constrained training.

\bibliography{custom}

\appendix

\section{Complete Derivation}
\label{app:derivations}
\subsection{Base Framework: Optimal Allocation}
\label{app:optimal_allocation}

We begin by deriving the intrinsic relationship between the optimal FLOPs ratio $r^*$ and the total compute $C$, assuming fixed architectural parameters. The model’s loss $\mathcal{L}$ is governed by two competing factors:

\paragraph*{Attention Capacity} ($\mathcal{A}$): This represents the ability to model token interactions and scales sublinearly with attention compute $C_A$. The reason for this sublinear scaling is that as the model's attention mechanism captures more long-range dependencies in the input sequence, the additional improvement from each extra unit of compute becomes progressively smaller. Beyond a certain point, the model reaches a level where further compute contributes less to the overall understanding of token interactions. This diminishing return is commonly observed in large-scale attention models, which can be captured by the relation $\mathcal{A}(C_A) = C_A^{\mu_A}$, where $\mu_A \in (0,1)$.

\paragraph*{Expert Capacity} ($\mathcal{E}$): This refers to the model's ability to process specialized features and scales sublinearly with expert compute $C_E$. As more compute is allocated to experts, the model’s performance improves at first, but the gains decrease due to the overhead of managing additional experts and the limited interactions between them. Furthermore, in MoE models, only a subset of experts are activated for each token, so the additional compute does not always result in proportionally more active experts, leading to diminishing returns. This behavior is reflected in the scaling law $\mathcal{E}(C_E) = C_E^{\mu_E}$, where $\mu_E \in (0,1)$.

The total loss $\mathcal{L}$ is modeled as a weighted sum of these two components:

\begin{equation}
    \mathcal{L} = \alpha_A \cdot \mathcal{A}(C_A)^{-\gamma_A} + \alpha_E \cdot \mathcal{E}(C_E)^{-\gamma_E},
\end{equation}

where $\gamma_A$ and $\gamma_E$ represent diminishing returns for attention and expert capacities, respectively, and $\alpha_A$ and $\alpha_E$ are architecture-dependent coefficients.

\subsection{Optimality Condition}

The goal is to find the optimal compute allocation between attention and expert capacity, defined by the ratio $r^* = C_E / C_A$, which minimizes the model’s loss. The optimal ratio is found by setting the derivatives of the loss with respect to $C_A$ and $C_E$ equal to each other:

$$
\frac{\partial \mathcal{L}}{\partial C_A} = \frac{\partial \mathcal{L}}{\partial C_E} \implies \frac{\alpha_A \gamma_A \mu_A}{C_A^{\gamma_A \mu_A + 1}} = \frac{\alpha_E \gamma_E \mu_E}{C_E^{\gamma_E \mu_E + 1}}.
$$

To proceed, we express $C_A$ and $C_E$ in terms of the total compute $C$ and the FLOPs ratio $r$. Assuming the following relationships:

$$
C_A = \frac{C}{1 + r}, \quad C_E = \frac{C r}{1 + r},
$$

we substitute these into the optimality condition and solve for $r^*$. This gives the optimal FLOPs ratio $r^*$ as:

\begin{equation}
    r^* = \alpha_r \cdot C^{\beta_r},
\end{equation}
where $\alpha_r$ and $\beta_r$ are defined as: 

$$
\alpha_r = \left( \frac{\alpha_E \gamma_E \mu_E}{\alpha_A \gamma_A \mu_A} \right)^{\frac{1}{\gamma_A \mu_A + \gamma_E \mu_E + 1}}, 
$$

$$
\quad \beta_r = \frac{\gamma_E \mu_E - \gamma_A \mu_A}{\gamma_A \mu_A + \gamma_E \mu_E + 1}.
$$

This relationship establishes a power-law scaling between the optimal ratio $r^*$ and total compute $C$, with the constants $\alpha$ and $\beta$ determined by the elasticity parameters $\mu_A, \mu_E, \gamma_A, \gamma_E$.

\section{Experimeal Setup}
\label{app:experimental_setup}

In this section, we detail the core components and procedures underpinning our empirical evaluation. We begin by describing the model architecture, including the design of sparse MoE layers and the configuration used to control per-token compute. Next, we outline the composition and sourcing of our training data, highlighting its multilingual and multimodal nature. Finally, we present our strategy for selecting the optimal FLOPs ratio \(r^*\), including criteria for handling unexpected fluctuations in the loss landscape. Together, these subsections provide a comprehensive overview of how we ensure fairness, reproducibility, and practical relevance in our experiments.

\subsection{Model Architecture and Training Configuration}
We implement a GPT-style decoder-only Transformer model, with sparse Mixture-of-Experts (MoE) layers replacing the conventional feed-forward networks (FFNs). Each MoE layer is designed to activate a subset of experts, specifically \textbf{3 experts per token}, consisting of 1 shared expert and 2 routed experts. This design yields a sparsity factor $S = \frac{E - 3}{E}$, where $E$ represents the total number of experts per layer. We evaluate models with varying expert counts, specifically $E \in \{17, 33, 65, 129\}$, which correspond to sparsity levels of $S \in \{82.35\%, 90.91\%, 95.38\%, 97.67\%\}$, covering a broad spectrum of sparsity commonly found in current MoE models. This range ensures that our findings are broadly applicable and relevant to contemporary MoE model designs.

To ensure fairness and consistency in our experiments, we maintain a constant total per-token compute budget across all models. The per-token compute cost, $C_{\text{token}}$, is defined as:

$$
C_{\text{token}} = \frac{C_A + C_E}{n_{\text{token}}},
$$
where $C_A$ and $C_E$ represent the compute costs for the attention and expert layers, respectively, and $n_{\text{token}}$ is the number of tokens processed. We vary the compute allocation ratio, $r = \frac{C_E}{C_A}$, in the range of $r \in [0.2, 1.5]$, adjusting the dimensions of the attention and expert layers while keeping the total compute budget fixed.


We conduct experiments across six benchmark models of different sizes—spanning a total parameter count $N$ from 100 M to 5 B, with the largest training FLOPs reaching $1 \times 10^{21}$—for each sparsity level. For each model, we fix the per-token compute budget $C_{\text{token}}$ and vary the FLOPs ratio $r$ by independently scaling the dimensions of the attention and expert sublayers. All variants share the same depth (number of layers) and number of attention heads. Each model is assigned the activation-parameter label $N_a$ based on its corresponding benchmark, ensuring that models with identical $C_{\text{token}}$ but different $r$ share the same label. Prior to experiments, we performed a hyperparameter search over learning rates and batch sizes for each label, finding that all models within a label shared a single optimal combination of learning rate and batch size. These hyperparameter settings are summarized in Table \ref{tab:hyperparams}. For consistency across all experiments, we fix the context length to $n_{\text{ctx}} = 4096$ tokens and the vocabulary size to $n_{\text{vocab}} = 128\text{k}$ tokens. All experiments were conducted on an A100 GPU cluster.

\begin{table*}[t]
\centering
\caption{Key Hyperparameters}
\label{tab:hyperparams}
\begin{tabular}{lcccc}
\toprule
Label & Layers ($n_\text{layer}$) & Heads ($n_\text{head}$) & Batch Size & Learning Rate\\
\midrule
20M & 8 & 8 & 96 & 0.0015\\ 
30M & 8 & 8 & 160 & 0.0013 \\ 
55M & 10 & 10 & 224 & 0.0011 \\ 
100M & 14 & 12 & 320 & 0.0009\\ 
200M & 16 & 16 & 512 & 0.0008 \\ 
\bottomrule
\end{tabular}
\end{table*}

\subsection{Datasets}
The training dataset is carefully curated to provide a comprehensive and balanced representation of diverse domains, ensuring the model's generalization across various languages and content types. The dataset is composed primarily of Chinese (15\%), English (60\%), and code (25\%), reflecting the model's focus on multilingual and multimodal capabilities. The content is sourced from a wide variety of domains, with the largest proportion coming from web-based data (50\%), followed by code (25\%), books (5\%), academic sources (5\%), and math-related content (5\%). Other domains, including news, social media, and exam-related content, contribute smaller yet significant portions.

This diverse selection of training data not only enables the model to learn from a wide range of linguistic and contextual patterns but also facilitates its ability to generate high-quality content across different modalities. By incorporating a mixture of textual and code-based data, the model is equipped to handle a variety of real-world tasks, from natural language understanding and generation to code completion and debugging. The inclusion of data from multiple languages further enhances its robustness, allowing for effective processing and generation in both English and non-English contexts.

\subsection{Selection Strategy for the Optimal Point $r^*$}

In our experimental setup, for a fixed set of parameters—specifically, the total FLOPs $C$ and sparsity $S$—the optimal value of $r$ is typically chosen based on the lowest loss observed during the experiments. This value of $r$ is selected as the optimal point $r^*$.

However, in cases where the actual optimal point does not align with expectations, we consider the possibility of selecting a suboptimal point. This occurs when, despite an increase in the total FLOPs, the value of the optimal ratio $r^*$ decreases. This observation contradicts the common industry expectation that, as training volume increases, the proportion of attention should decrease. In such instances, we interpret this behavior as a fluctuation, rather than a clear deviation. Consequently, we select the suboptimal point as the theoretical optimal point, provided that the loss difference between the suboptimal point and the actual optimal point is less than 0.001.

\section{Estimating MoE FLOPs}

In this section, we describe the computation of floating-point operations (FLOPs) for training a Mixture of Experts (MoE) model, following \citet{narayanan2021efficient}. Specifically, we consider the forward and backward passes, with a focus on matrix multiplications that dominate the computational cost. This section breaks down the FLOPs into several key components: \textbf{attention mechanisms}, \textbf{sparse MLP layers}, and \textbf{output layers}. The FLOPs for the backward pass are assumed to be approximately twice that of the forward pass unless otherwise specified.

\subsection{Attention Mechanism FLOPs}

The attention mechanism in a Transformer model consists of several stages, including query, key, and value projections, followed by the computation of the attention weights and their application to the values.

\paragraph{Query Projection} The query projection is computed as a matrix multiplication between the input sequence and the query projection matrix. The FLOPs for the query projection are given by:

   $$
   C^\text{Q}_\text{proj} = 2n_\text{ctx}d_\text{hidden}^2.
   $$

\paragraph{Key-Value Projection} For MoE models, the number of key and value heads is typically different from the number of attention heads. If a generic query architecture (GQA) is used, the key and value projections require separate computations for the keys and values. The corresponding FLOPs are given by:

   $$
   C^\text{KV}_\text{proj} = 4n_\text{ctx}d_\text{hidden}^2\div \text{kv head ratio}.
   $$

   Otherwise, without GQA, the key-value projection FLOPs are twice that of the query projection:

   $$
   C^\text{KV}_\text{proj} = 2C^\text{Q}_\text{proj}.
   $$

\paragraph{Attention Weights and Value Application}The computation of the attention weights involves matrix multiplications between the queries and keys, and their application to the values:

   $$
   C^\text{Attn}_\text{weight} = 2n_\text{ctx}^2 d_\text{hidden}.
   $$

   $$
   C^\text{Value} = 2n_\text{ctx}d_\text{hidden}^2.
   $$

\paragraph{Final Output Projection}After applying the attention weights, a final output projection is computed:

   $$
   C^\text{Output}_\text{proj} = 2n_\text{ctx}d_\text{hidden}^2.
   $$

The total FLOPs for the attention mechanism (including both projections and attention computations) are then given by:

\begin{equation*}
    \begin{aligned}
        \text{Attn Forward FLOPs}  =  &\text{Attn Projection FLOPs} + \\
&\text{Attn Mechanism FLOPs}.
    \end{aligned}
\end{equation*}

\subsection{Sparse MLP Forward FLOPs}

In MoE models, the sparse expert mechanism introduces additional computational overhead due to the dynamic selection of experts for each token. The sparse MLP forward pass involves a combination of linear transformations and matrix multiplications across the selected experts.

The FLOPs for the sparse MLP forward pass are calculated as follows:

\begin{equation*}
\begin{aligned}
C_{\text{expert}}
&= n_{\text{ctx}} \cdot \Bigl(
    2 d_{\text{hidden}} E
    + 3 \cdot 2 \cdot d_{\text{hidden}} d_{\text{expert}} \\
&\qquad\qquad\qquad \cdot \bigl(\text{top-}k + n_{\text{shared\_experts}}\bigr).
\Bigr).
\end{aligned}
\end{equation*}

\subsection{Output Layer FLOPs}

Finally, the output logits layer computes the probability distribution over the vocabulary for each token. The FLOPs for the output layer are given by:

$$
C_\text{logits} = 2n_\text{ctx}d_\text{hidden}n_\text{vocab}.
$$

\subsection{Decoder Layer FLOPs}

Each transformer decoder layer combines attention and sparse MLP computations. The total FLOPs for a single decoder layer forward pass are therefore the sum of the attention and sparse MLP FLOPs:

\begin{equation*}
    \begin{aligned}
        C^\text{layer}_\text{forward} & = \text{Attention Forward FLOPs} \\
        & + \text{Sparse MLP Forward FLOPs}.
    \end{aligned}
\end{equation*}

\paragraph{Total Forward Pass FLOPs}

The total forward pass FLOPs for the entire model, excluding gradient checkpointing, is given by:

$$
C_\text{forward} = n_\text{layer}C^\text{layer}_\text{forward} + C_\text{logits}.
$$

\paragraph{Backward Pass FLOPs}

The backward pass typically incurs twice the computational cost of the forward pass, as it requires the computation of gradients for all parameters. If PEFT (Parameter-Efficient Fine-Tuning) is used, we assume that the backward pass will require a factor of 2. Otherwise, we use a factor of 3:

$$
C_\text{backward} = C_\text{forward} \cdot \text{factor}.
$$

\paragraph{Gradient Checkpointing}

If gradient checkpointing is enabled, the forward pass computations are adjusted to store intermediate results during the forward pass to reduce memory usage, but at the cost of additional computational overhead. The total FLOPs in this case are given by:

$$
C_\text{forward} = n_\text{layer}C^\text{layer}_\text{forward}\cdot (\text{factor} + 1) + C_\text{logits} \cdot \text{factor}.
$$

Given these calculations, the total hardware FLOPs and model FLOPs can be computed for both the forward and backward passes. The specific computational cost depends on factors such as the use of gradient checkpointing, the PEFT setting, and the number of active experts in the MoE model.

\section{Detailed Analysis and Visualizations of the Scaling Law Fitting Process}

\subsection{Initialization Parameters for L-BFGS Optimization}

Table~\ref{tab:initial_values} shows the parameters used to initialize L-BFGS, which was employed to fit the proposed parametric scaling law presented in Equation~\ref{eq:final}.
\begin{table*}[t]
\centering
\begin{tabular}{@{}lc@{}}
\toprule
\textbf{Coefficients} & \textbf{Initial Values} \\ \midrule
$\log(a), \log(b), \log(c), \log(d)$ & [0, 10, 20] \\
$\alpha, \beta, \lambda, \gamma$ & [0, 0.25, 0.5, 0.75, 1, 1.25] \\
$\log(\tau)$ & 1.5 \\ \bottomrule
\end{tabular}
\caption{Initial values used to estimate coefficients in Equation~\ref{eq:final}.}\label{tab:initial_values}
\end{table*}
\subsection{Justification for the Efficiency Loss Term}
\label{app:mathematical_properties}
To justify the use of the $\frac{r}{r+1}$ form for the efficiency loss term, we analyze its mathematical properties in relation to diminishing returns. This function exhibits three key characteristics that align with empirical observations of expert compute allocation:
\begin{itemize}
    \item \textbf{Bounded Growth}: As $r \to \infty$, $\frac{r}{r+1} \to 1$, preventing unbounded loss growth while maintaining sensitivity to allocation changes in practical $r$ ranges (typically $r \in [0.2, 1.5]$ in our experiments). This reflects the physical reality that over-allocating compute to experts cannot infinitely degrade performance.
    \item \textbf{Sublinear Scaling}: The derivative $\frac{d}{dr}\left(\frac{r}{r+1}\right) = \frac{1}{(r+1)^2}$ decreases quadratically with $r$, explicitly modeling diminishing marginal returns from increasing expert allocation. This matches our empirical finding that performance gains from expert compute saturate faster than those from attention compute.
    \item \textbf{Zero-Cost Baseline}: At $r=0$ (no expert allocation), the term vanishes ($\frac{0}{0+1}=0$), ensuring no artificial penalty when experts are disabled. The linear regime $\frac{r}{r+1} \approx r$ when $r \ll 1$ matches observed near-linear improvements at small $r$ values.
\end{itemize}

Alternative formulations like $\log(1+r)$ or $1-e^{-r}$ were tested but showed poorer empirical fit, particularly in matching the saturation behavior observed at high $r$. The chosen form provides the best trade-off between empirical accuracy and parameter efficiency, requiring only a single scaling coefficient $d$ to capture the full efficiency loss trajectory.

\subsection{Visualizations of Alternative Formula Fits}

In this subsection, we compare the observed validation losses with the predictions made by two previously proposed scaling formulas, using scatter plots to assess their fitting quality.

\begin{figure*}[t]
    \centering
    \subfigure[\citet{wang-etal-2024-scaling}]{\label{subfigure:wang}
        \includegraphics[width=0.48\linewidth]{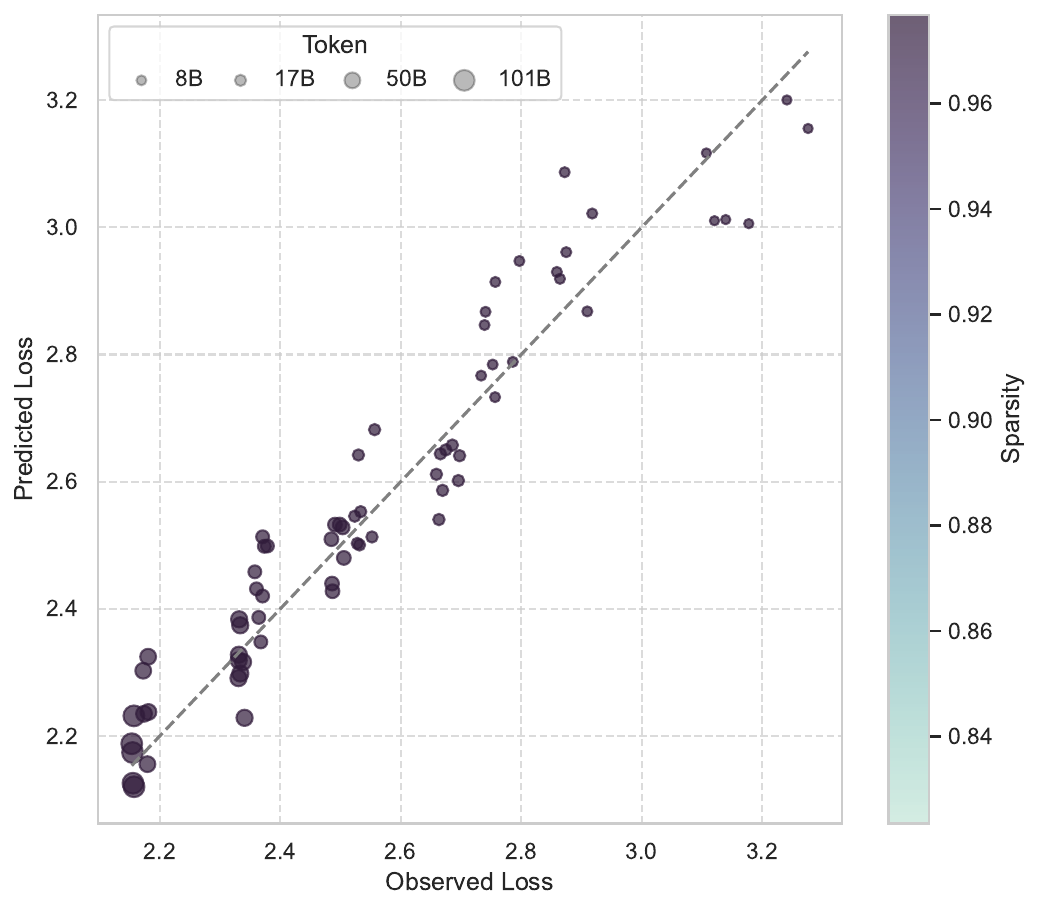}
    }
    \subfigure[\citet{abnar2025parameters}]{\label{subfigure:apple}
        \includegraphics[width=0.48\linewidth]{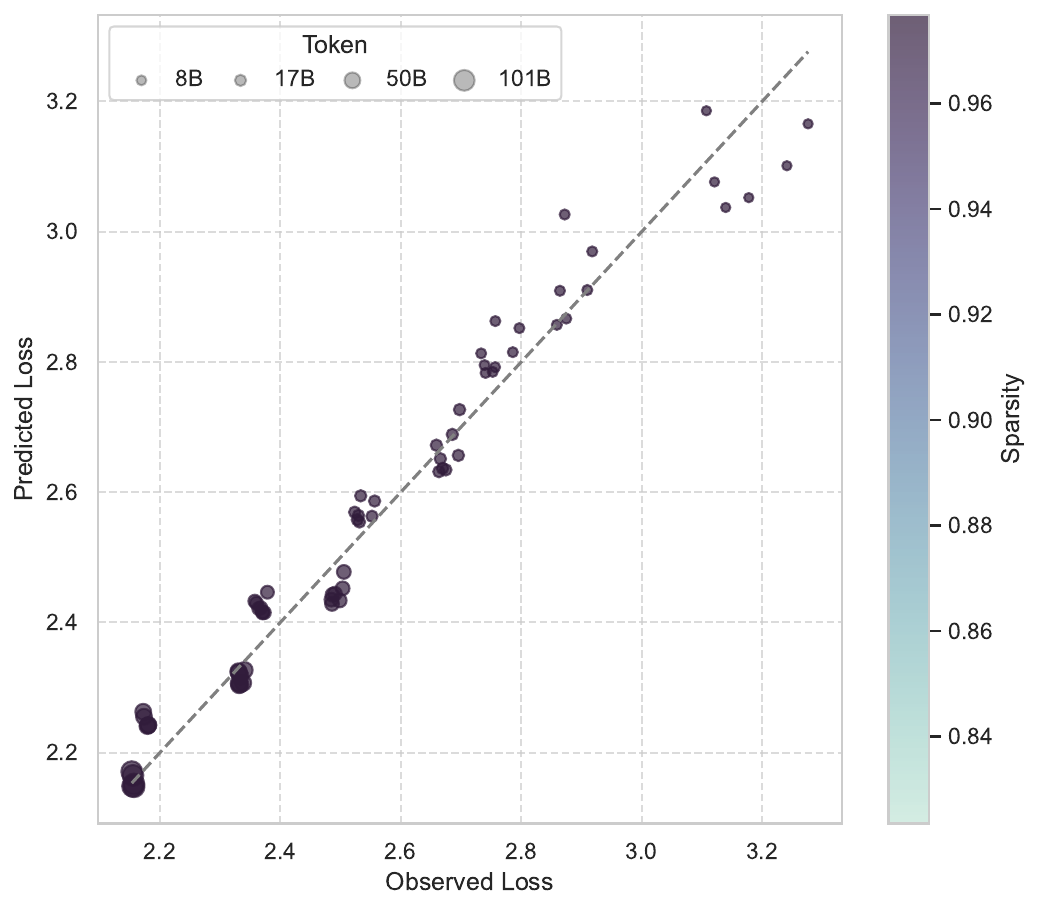}
    }
    \caption{Comparison of observed versus predicted validation losses for two alternative scaling formulas. (a) Predictions using the \citet{wang-etal-2024-scaling} formulation (Equation~\ref{eq:wang}), which fails to generalize across modern high-sparsity settings. (b) Predictions using the \citet{wang-etal-2024-scaling} formulation, which fits large models well but underperforms on smaller ones. The solid diagonal line indicates perfect agreement.}
    \label{fig:val_comparison}
\end{figure*}

\citet{wang-etal-2024-scaling} extend the traditional dense-scaling law by introducing the total number of experts \(E\) as an additional factor:
\begin{equation}\label{eq:wang}
    \mathcal{L} = \frac{a}{N^\alpha E^\gamma} + \frac{b}{D^\beta} + \tau.
\end{equation}
However, this formulation ignores the fraction of activated experts and is constrained to \(E < 100\), which limits its applicability to modern high-sparsity, fine-grained MoE models \citep{deepseekai2024deepseekv3technicalreport}. Figure~\ref{subfigure:wang} shows that, across all model scales, Equation~\ref{eq:wang} fails to capture the true loss trends.

\citet{abnar2025parameters} incorporate sparsity \(S\) into their fit:
\begin{equation}
    \mathcal{L} = \frac{a}{N^\alpha} + \frac{b}{N^\beta} + \frac{c}{(1-S)^\gamma}
                   + \frac{d}{(1-S)^\delta \, N^\gamma} + \tau.
\end{equation}
As plotted in Figure~\ref{subfigure:apple}, this formula performs reasonably well on large models but still exhibits substantial errors for smaller ones.

\section{Additional Visualizations}
\label{app:addtional_visual}

\begin{figure*}[t]
    \centering
    \subfigure{\includegraphics[width=0.48\linewidth]{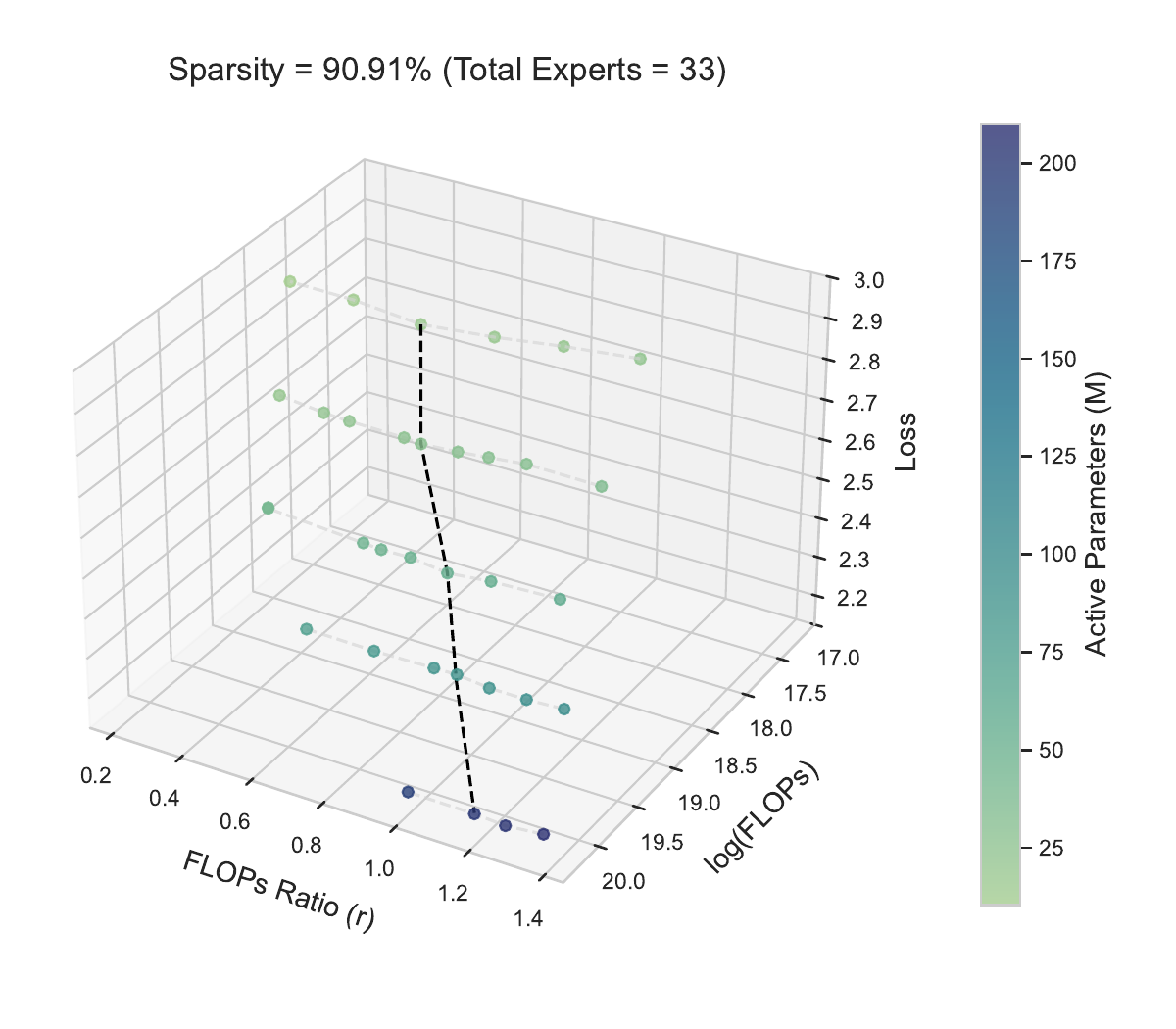}}
    \subfigure{\includegraphics[width=0.48\linewidth]{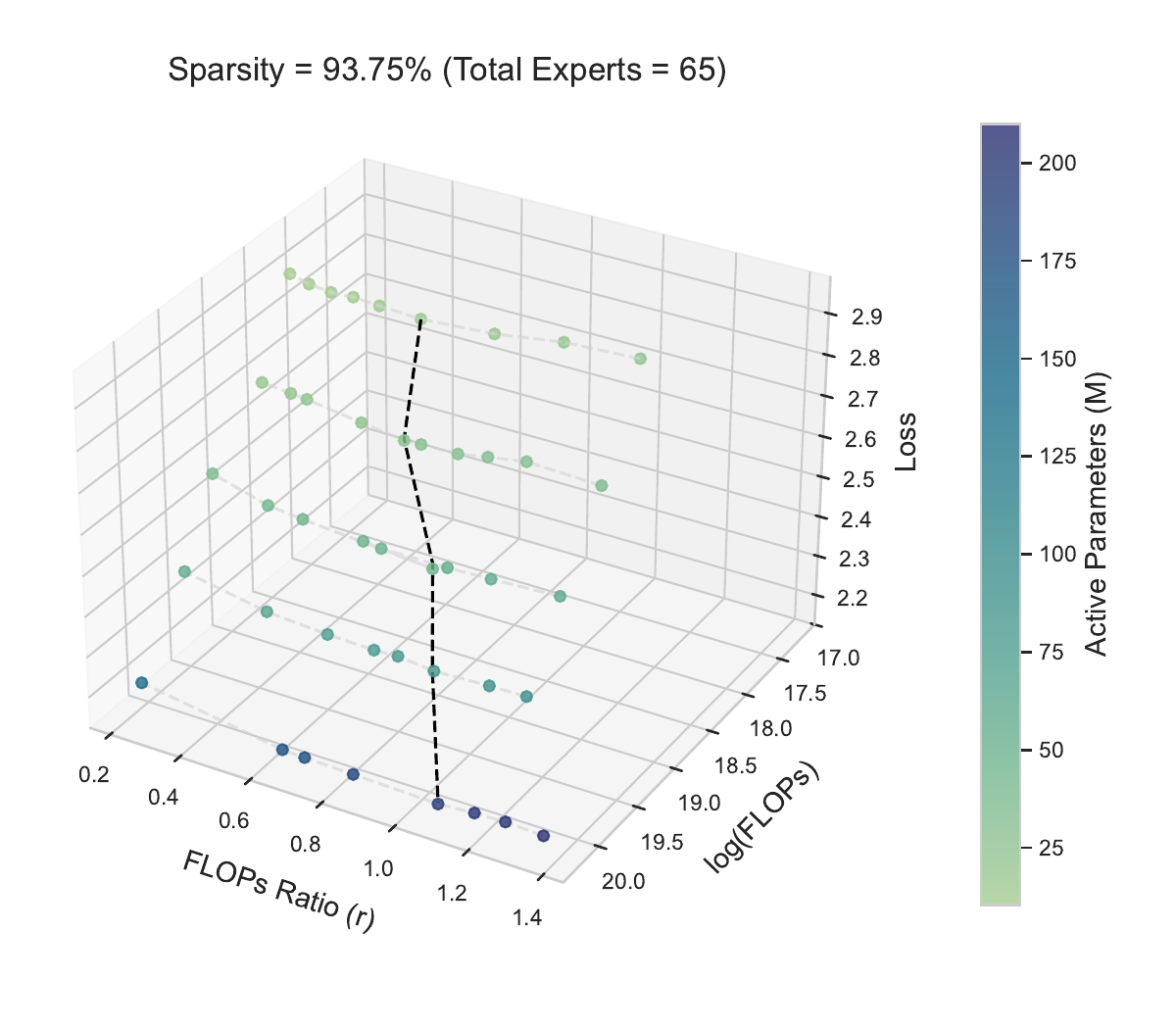}}
    \caption{Loss as a function of FLOPs ratio and total compute. Black dashed lines trace optimal $r^*$. Color indicates active parameters. Low-sparsity models (left) favor higher $r^*$ at scale.}
    \label{fig:flops_ratio_loss_appendix}
\end{figure*}

\end{document}